\documentclass[review]{elsarticle}

\usepackage{amsmath}
\usepackage{amssymb}

\journal{Journal of \LaTeX\ Templates}

\begin{document}

\begin{frontmatter}

\title{Unpaired Photo-to-Caricature Translation on Faces in the Wild}




\author[oucee]{Ziqiang Zheng}
\author[oucee]{Chao Wang}
\author[oucee]{Zhibin Yu}
\author[oucee]{Nan Wang}
\author[oucee]{Haiyong Zheng\corref{correfauthor}}
\cortext[correfauthor]{Corresponding author}
\ead{zhenghaiyong@ouc.edu.cn}
\author[oucee]{Bing Zheng}
\address[oucee]{No. 238 Songling Road, Department of Electronic Engineering, Ocean University of China, Qingdao, China}

\begin{abstract}
Recently, image-to-image translation has been made much progress owing to the success of conditional Generative Adversarial Networks (cGANs). And some unpaired methods based on cycle consistency loss such as DualGAN, CycleGAN and DiscoGAN are really popular. However, it's still very challenging for translation tasks with the requirement of high-level visual information conversion, such as photo-to-caricature translation that requires satire, exaggeration, lifelikeness and artistry. We present an approach for learning to translate faces in the wild from the source photo domain to the target caricature domain with different styles, which can also be used for other high-level image-to-image translation tasks. In order to capture global structure with local statistics while translation, we design a dual pathway model with one coarse discriminator and one fine discriminator. For generator, we provide one extra perceptual loss in association with adversarial loss and cycle consistency loss to achieve representation learning for two different domains. Also the style can be learned by the auxiliary noise input. Experiments on photo-to-caricature translation of faces in the wild show considerable performance gain of our proposed method over state-of-the-art translation methods as well as its potential real applications.
\end{abstract}


\end{frontmatter}


\section{Introduction}

Image-to-image translation has been made much progress~\cite{Isola_2017_CVPR,Zhu_2017_ICCV,Yi_2017_ICCV,Chen_2017_ICCV,kim2017learning,mao2018semantic} because many tasks~\cite{zhou2016similarity,wang2016super,ma2017unsupervised,shi2017end} in image processing, computer graphics, and computer vision can be posed as translating an input image into a corresponding output image~\cite{zhu2016generative,li2016precomputed,wang2016generative,Ledig_2017_CVPR,taigman2017unsupervised,Sela_2017_ICCV,Tung_2017_ICCV}. And its achievements mainly owes to the success of Generative Adversarial Networks (GANs)~\cite{goodfellow2014generative}, especially conditional GANs (cGANs)~\cite{mirza2014conditional,radford2016unsupervised,Isola_2017_CVPR}. However, the current studies mainly concern image-to-image translation tasks with low-level visual information conversion, \emph{e.g.},  photo-to-sketch~\cite{liu2018auto}.

A caricature is a rendered image showing the features of its subject in an exaggerated way and usually used to describe a politician or movie star for political or entertainment purpose. Creating caricatures can be considered as artistic creation tracing back to the 17th century with the profession of caricaturist. Then some efforts have been made to produce caricatures semi-automatically using computer graphics techniques~\cite{akleman2000making}, which intend to provide warping tools specifically designed toward rapidly producing caricatures. But there are very few software programs designed specifically for automatically creating caricatures, and to the best of our knowledge, none can work to be comparable with caricaturist. Nowadays, besides the political and public-figure satire, caricatures are also used as gifts or souvenirs, and more and more museums dedicated to caricature throughout the world were opened. So it would be very useful and meaningful if computers can create caricatures from photos automatically and intelligently.

Photo-to-caricature is a typical high-level image-to-image translation problem but with bigger challenge than other low-level translation problems such as photo-to-label, photo-to-map, or photo-to-sketch~\cite{Isola_2017_CVPR}, because caricatures
\begin{itemize}
\item require \textbf{satire} and \textbf{exaggeration} of photos;
\item need \textbf{artistry} with different styles;
\item must be \textbf{lifelike}, especially the expression of a face photo.
\end{itemize}
Specifically, for a face photo, we want to create the face caricatures with different styles, which exaggerate the face shape or facial sense organs (\emph{i.e.}, ears, mouth, nose, eyes and eyebrow) but keep the vivid expression while producing artistry.

\begin{figure}[!ht]
\centering
\includegraphics[width=\linewidth]{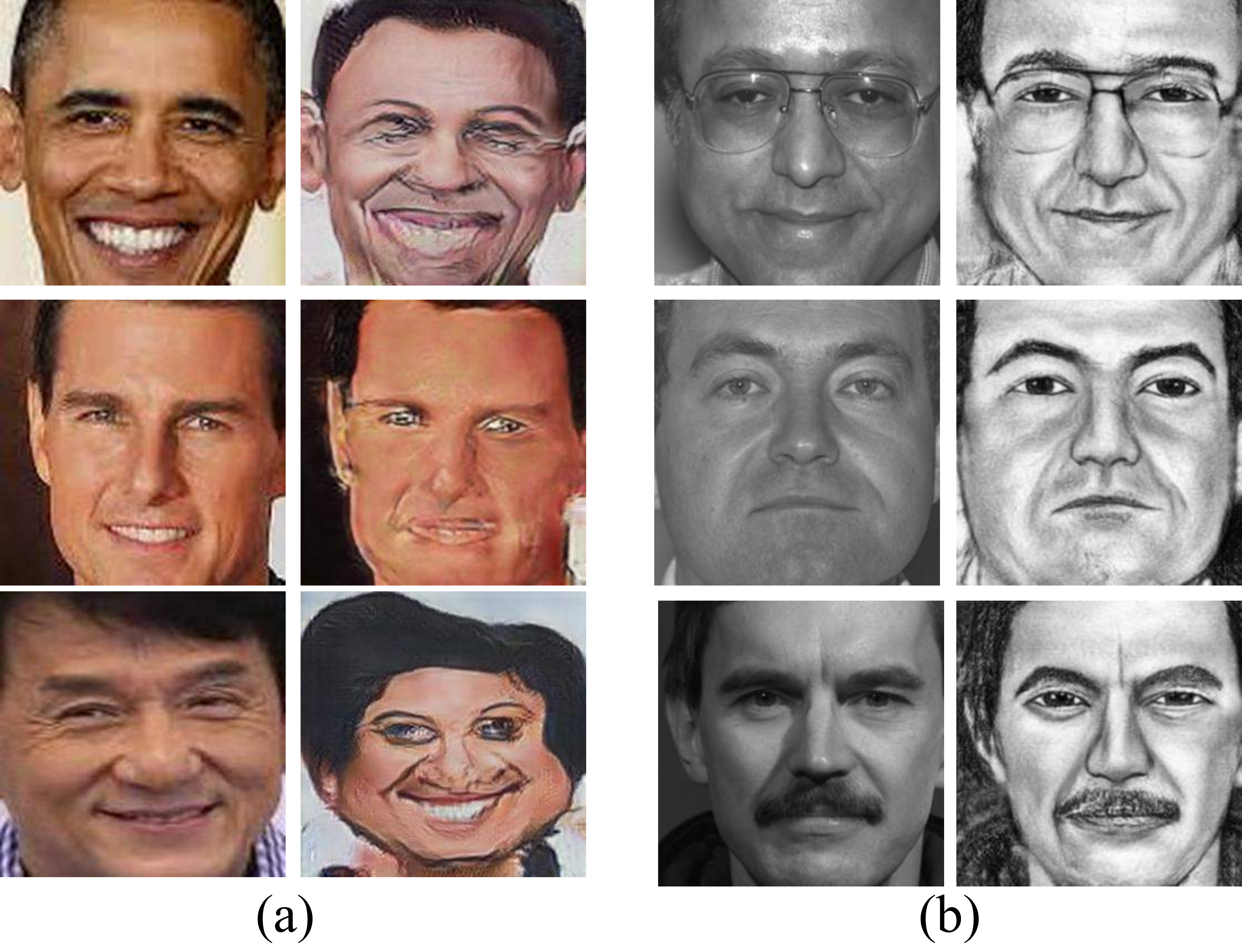}
\caption{Translating faces in the wild from photo to caricature with different styles by our proposed method. (a) Example results on IIIT-CFW dataset~\cite{mishra2016iiit}; (b) Example results on PHOTO-SKETCH dataset~\cite{zhang2011coupled,wang2009face}.}
\label{fig:intro}
\end{figure}

In this paper, we propose a GAN-based method for learning to translate faces in the wild from the source photo domain to the target caricature domain (see Figure~\ref{fig:intro} for translating examples and Figure~\ref{fig:p2c} for the architecture of our proposed method). Although deep convolutional neural networks with adversarial training~\cite{radford2015unsupervised} can generate images with enough precise facial features~\cite{berthelot2017began}, these images sometimes still have wrong relationships between facial features or mismatch among facial organs such as nose and eyes, \emph{e.g.}, a face with more than two eyes or crooked nose. Traditional GANs can produce correct facial organs but wrong relationships between them, and also it's very challenging to abstract and exaggerate face and facial organs. We attribute this problem to the deficient capacity of the discriminator of GAN to distinguish real-fake images. Therefore, our motivation is to design an adversarial training with multiple discriminators to improve the ability of GAN's discriminator for feature representation.

Based on the model of CycleGAN~\cite{Zhu_2017_ICCV}, we design a dual pathway model of GAN for high-level image-to-image translation tasks, where one pathway of coarse discriminator is in charge of abstracting the global structure information, while another pathway of fine discriminator is responsible for concerning the local statistics information. For generator, besides the adversarial loss, we provide one more extra perceptual similarity loss to constrain consistency for generated output itself and with the unpaired target domain image. By using our proposed method, the photos of faces in the wild can be translated to caricatures with learned general-purpose exaggerated artistic styles while still keeping the original lifelike expression (see Figure~\ref{fig:intro} and \ref{fig:free} for references). Considering that traditional GANs are not robust and easily attracted by noise, we design a noise-added training procedure to improve the robustness of our model. Inspired by InfoGAN~\cite{chen2016infogan}, we find that auxiliary noise can help model learning the caricature style information while translating images in our task.

We have extensively evaluated our method on IIIT-CFW dataset~\cite{mishra2016iiit}, PHOTO-SKETCH dataset~\cite{zhang2011coupled,wang2009face}, Caricature~\cite{abaci2015matching}, FEI dataset~\cite{thomaz2010new}, Yale dataset~\cite{georghiades1997yale}, KDEF dataset~\cite{lundqvist1998karolinska} and CelebA dataset~\cite{liu2015faceattributes}. The experimental results show that our method can create acceptable caricatures from face photos while current state-of-the-art image-to-image translation methods can't. Also the designed experiments indicate the effectiveness of our proposed dual pathway of discriminators, additional noise input and extra perceptual loss, respectively. Besides, we tested our photo-to-caricature translation method for producing caricatures with adding different proportions of noise to show the translating robustness and style diversity. Furthermore, the proposed method can create caricatures for arbitrary face photos without pre-training on extra face datasets. Another prominent performance of our methods is that our model can capture the expression information and make some abstraction and exaggeration. This might be helpful to fill aforementioned gap of automatic and intelligent caricature creation.

\section{Related Work}

\textbf{Image-to-image translation.} Owing to the success of GANs~\cite{goodfellow2014generative}, especially various conditional GANs~\cite{mirza2014conditional,mathieu2015deep,reed2016generative,yan2016attribute2image,wang2016generative,zhang2017image}, image-to-image translation problems have been made much progress recently, which aims to translate an input image from one domain to another domain given input-output images pairs~\cite{Isola_2017_CVPR}. Earlier image-conditional models for specific applications have achieved impressive results on inpainting~\cite{Pathak_2016_CVPR}, de-raining~\cite{zhang2017image}, texture synthesis~\cite{li2016precomputed}, style transfer~\cite{wang2016generative}, video prediction~\cite{mathieu2015deep} and super-resolution~\cite{Ledig_2017_CVPR}. The general-purpose solution for image-to-image translation developed by Isola \emph{et al.}~\cite{Isola_2017_CVPR} with the released \verb|Pix2pix| software has achieved reasonable results on many translation tasks by using paired images for training such as photo-to-label, photo-to-map and photo-to-sketch. Then CycleGAN~\cite{Zhu_2017_ICCV}, DualGAN~\cite{Yi_2017_ICCV}, and DiscoGAN~\cite{kim2017learning} were proposed for unpaired or unsupervised image-to-image translation with almost comparable results to paired or supervised methods. However, the translation tasks that these work can tackle usually concern the conversion of low-level visual information such as line (photo$\rightarrow$sketch), color (summer$\rightarrow$winter), and texture (horse$\rightarrow$zebra), but it's still very challenging for some translation tasks with the requirement of high-level visual information conversion, \emph{e.g.}, abstraction and exaggeration (photo$\rightarrow$caricature).
Recently, Iizuka \emph{et al.}~\cite{iizuka2017globally} combined two discriminators called global discriminator and local discriminator to improve the adversarial training for image completion task. Experimental results have proven that the global discriminator can distinguish the images based on the global parts while the local discriminator pays attention to the details of parts. We exploit the design of two discriminators for high-level image-to-image translation tasks in this paper.

\textbf{Photo-to-cartoon translation.} Translating photo to cartoon by computer algorithms has been studied for a long time because of the corresponding interesting and meaningful applications. The earlier work mainly relied on the analysis of facial features~\cite{luo2002exaggeration,chen2004automatic,liao2004automatic}, which is hard to be applicable for large-scale faces in the wild. Thanks to the invention of GANs~\cite{goodfellow2014generative}, automatic photo-to-cartoon translation becomes feasible. But most of the current related work mainly focused on the generation of anime~\cite{zhang2017style} or emoji~\cite{taigman2017unsupervised} with specific style\footnote{We consider that anime, emoji, and caricature are three types of cartoon or three subsets of cartoon.}. And these works have nothing to do with caricature creation that needs to be exaggerated, lifelike and artistic.

Unlike the prior works for image-to-image translation dealing with low-level visual information conversion, our study mainly focuses on the translation problems of high-level visual information conversion, \emph{e.g.}, photo-to-caricature. For this purpose, our method differs from the past work in network architecture as well as the layers and losses. We design a dual pathway model of GAN with two discriminators named coarse discriminator and fine discriminator to capture global structure and local statistics respectively, and we apply the perceptual similarity loss for generator. For learning the style information and improving the robustness of our model, we provide the input with auxiliary noise. Here we show that our approach is effective on the face photo-to-caricature translation task, which typically requires high-level visual information conversion.

\section{Method}

\begin{figure}[!ht]
\centering
\includegraphics[width=\textwidth]{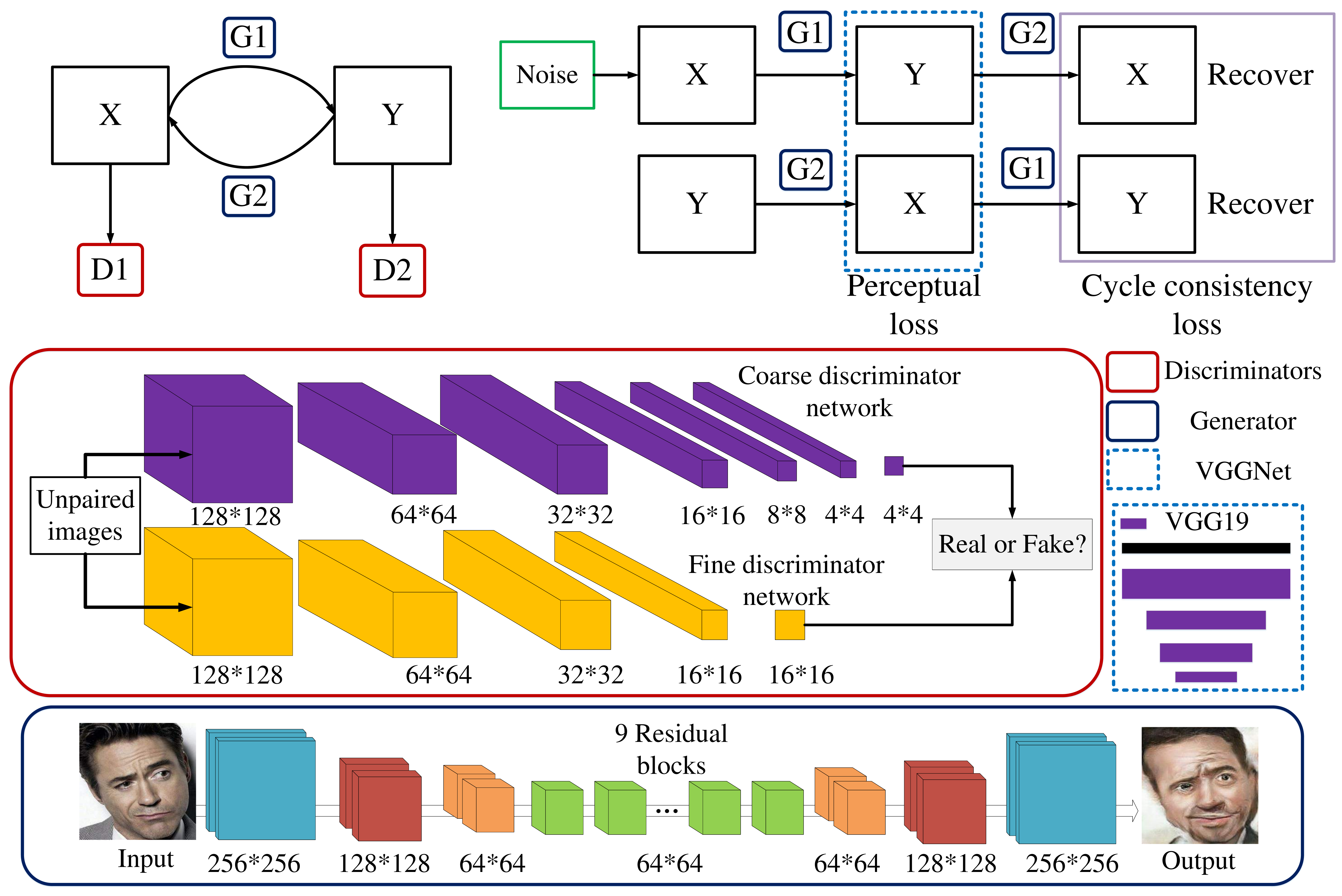}
\caption{Network architecture and data flow chart of our proposed method for face photo-to-caricature translation.}
\label{fig:p2c}
\end{figure}

Conditional GANs are generative models that learn a mapping from observed input image $x$, to target image $y$, $G:\left\{x\right\}\rightarrow y$. The objective of our conditional GAN can be expressed as
\begin{equation}
\begin{split}
\mathcal{L}_{c}(G,D)=\mathbb{E}_{x,y\sim P_{data}(x,y)}\left[\log D(x,y)\right]+\\
\mathbb{E}_{x\sim P_{data}(x)}\left[\log (1-D(x,G(x)))\right].
\end{split}
\label{eq:cgan}
\end{equation}
The goal is to learn a generator distribution over data $y$ that matches the real data distribution $P_{data}$ by transforming an observed input image $x\sim P_{data}(x)$ into a sample $G(x)$. This generator is trained by playing against an adversarial discriminator $D$ that aims to distinguish between samples from the true data distribution and the generator's distribution. To deal with the high-level visual information conversion for some more challenging image-to-image translation tasks, we exploit one discriminator to dual discriminators and add two more losses (perceptual loss and cycle consistency loss) to adversarial loss, so our method can tackle conversions of both the low-level (line, color, texture, \emph{etc.}) and the high-level (expression, exaggeration, artistry, \emph{etc.}) visual information. And in order to improve the robustness of our model, we provide our model a noise-added training and use auxiliary noise to learn the style information while translating. The overall network architecture and data flow are illustrated in Figure~\ref{fig:p2c}.

\subsection{Cycle consistency loss}
Researchers apply adversarial training to learn the mapping function between two different image domains. Here we use two same generators (named \verb|G1| and \verb|G2| in Figure~\ref{fig:p2c}) for observing the sample distribution of two domains. And $X$ denotes the photo domain while $Y$ denotes the caricature domain in our task. We use our generators to emulate the translation between $X$ and $Y$. However, only using adversarial loss can't guarantee a plausible generation. So researchers use Cycle Consistency Loss ($\mathcal{L}_{cyc}$) to help establishing mapping function between domains~\cite{Zhu_2017_ICCV}. And $\mathcal{L}_{cyc}$ is expressed as
\begin{equation}
\begin{split}
\mathcal{L}_{cyc}(G1,G2)=\mathbb{E}_{x\sim P_{data}(x)}\left[||G_{2}(G_{1}(x))-x||_{1}\right]\\
+\mathbb{E}_{y\sim P_{data}(y)}\left[||G_{1}(G_{2}(y))-y||_{1}\right],
\end{split}
\end{equation}
where $G_{1}$ and $G_{2}$ represent the two generators, and $x$ and $y$ are samples from $X$ and $Y$ domain respectively. We use $L1$ loss for the cycle consistency loss following Zhu \emph{et al.}~\cite{Zhu_2017_ICCV}.

\subsection{Perceptual loss}
\label{sec:percep}
To further reduce the space gap of possible mapping functions between domains, we apply the perceptual loss $\mathcal{L}_{p}$ for our model. For a constrained translation problem, finding an appropriate loss function is critical. We adopt the content loss of Gatys \emph{et al.}~\cite{gatys2016image}, which is also referred as a perceptual similarity loss or feature matching~\cite{bruna2015super,dosovitskiy2016generating,johnson2016perceptual,ledig2016photo}. We apply the perceptual loss to our model followed by the cycle consistence loss, and compute the perceptual loss between unpaired images from different domains to push generator to capture the feature representations. Let $\phi$ denotes a pre-trained visual perception network (we use pre-trained \verb|VGG19| in our experiments) and $n$ denotes the number of feature maps. Different layers in the networks represent low-to-high level information: from edges and color to object and semantic representation. Matching both low and high layers in the perception network can help achieving fantastic translation. And the perceptual loss can be expressed as
\begin{equation}
\mathcal{L}_{p}=\sum^{N}_{n}{\lambda}_{n}||{\Phi}_{n}(y)-{\Phi}_{n}(G(x);\theta))||_{1},
\end{equation}
where $y$ denotes the image from caricature domain in our task and $G(x)$ denotes the synthesized caricature image using $x$. Note that these two images are unpaired. 

\subsection{Auxiliary noise input}
\begin{figure}[!ht]
\centering
\includegraphics[width=0.8\textwidth]{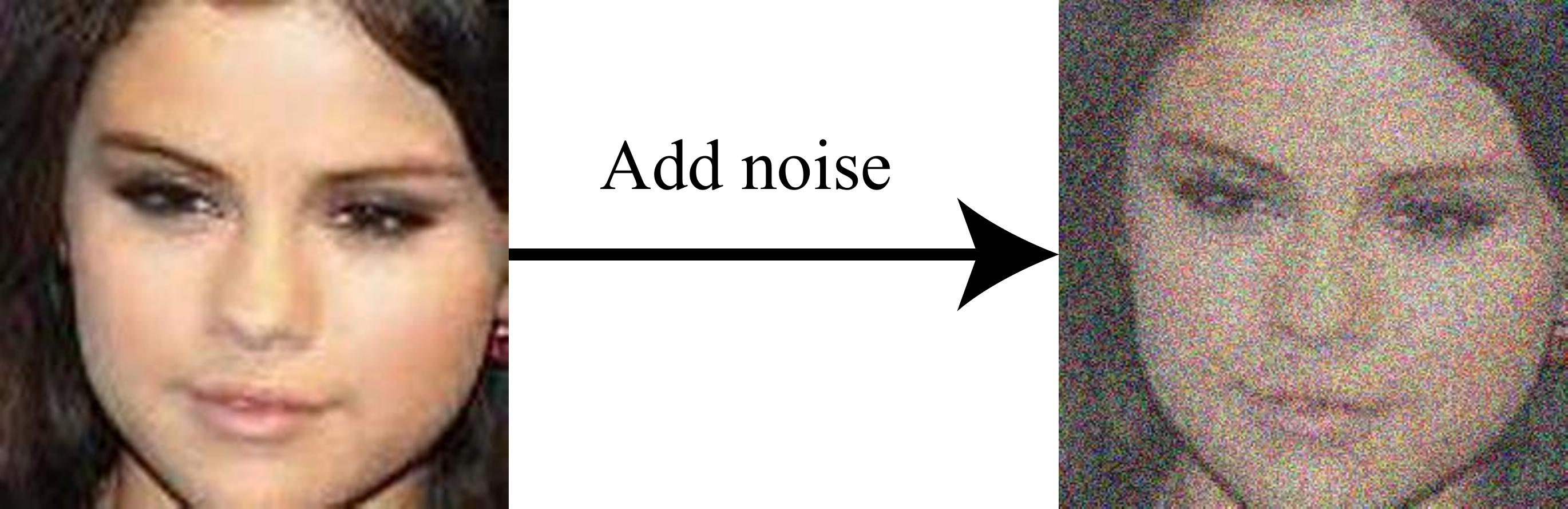}
\caption{Sample for adding noise, here we use $\alpha=0.5$.}
\label{fig:noise}
\end{figure}

Previous researches have fully proved that we can get plausible image results from noise input~\cite{radford2016unsupervised,chen2016infogan,zhao2016energy,berthelot2017began}. In order to improve the robustness and enrich the diversity of image translation between domains, we design a noise-added training procedure before the translation shown in Figure~\ref{fig:p2c}. First we obtain a random noise input from random uniform distribution (range from $0$ to $255$), then we merge the noise input and the raw image input to acquire the final input using approximate weights. Here we define $\alpha$ to denote the proportion of the raw image accounting for the final image. This can be expressed as
\begin{equation}
\label{eq:noise}
x=x_{i}*\alpha+(1-\alpha)*n, \quad 0\leq\alpha\leq1,
\end{equation}
where $x_{i}$ denotes the raw input from photo domain and $n$ denotes the noise that has a uniform distribution $P_{noise}$. Figure~\ref{fig:noise} shows one sample with adding noise. With auxiliary noise input, the GAN object is expressed as:
\begin{equation}
\begin{split}
L_{c}(G,D)=\mathbb{E}_{x,y\sim P_{data}(x,y)}\left[\log D_{p}\left(x,y\right)\right] \\
+ \mathbb{E}_{x\sim P_{data}(x),n\sim P_{noise}(n)}\left[\log (1-D(x,G(x,n)) \right].
\end{split}
\end{equation}

We add the auxiliary noise to improve the robustness of our model. And we can get different styles of output through adjusting the noise input (see Section~\ref{sec:info}). Besides, we add the noise from a uniform distribution for making the style information more representable and matching the style space while translating. Furthermore, we also hope to control the style by adding different noise and make the translating conditional, which will be our future work.

\subsection{Dual discriminators}
Traditional GANs usually have one generator and one discriminator to leverage adversarial training. Different from them, we design two different discriminators to capture different level information. 

In our method, we propose two different discriminators called coarse discriminator and fine discriminator respectively. The coarse discriminator aims to encourage generator to synthesize images based on global style and structure information for domain translation. In our high level image-to-image task, the coarse discriminator is capable for capturing the structure information and abstracting the representative information of face photo such as emotion and style. While the fine discriminator aims to achieve the feature matching and help generating more plausible and precise images, and the fine discriminator builds the adversarial training for the face details with generator such as lip and eyes. Different from Satoshi \emph{et al.}~\cite{iizuka2017globally}, we are not using the image patches as input of local discriminator, we provide both the two discriminators with the whole image as input while the outputs of the two discriminators are different (see Figure~\ref{fig:p2c}). The output of coarse discriminator is a $4\times 4$ patch matrix after the sigmoid activity function while the output of fine discriminator is $16\times 16$. Note that both the two discriminators are using sigmoid function at the last layer. We have tried using different output size combinations to get different results, and the experiments show that the combination of $16\times 16$ for fine discriminator and $4\times 4$ for coarse discriminator can obtain the best result for translation. The coarse discriminator has a smaller feature map with more abstractive representation compared to the fine discriminator. The \verb|D1| and \verb|D2| in Figure~\ref{fig:p2c} represent the dual discriminators for translating two different domains $X$ and $Y$. 

\subsection{Generator}
\label{sec:generator}
Previous studies have found it beneficial to mix the GAN objective with a more traditional norm, such as $L2$~\cite{Pathak_2016_CVPR} and $L1$~\cite{Isola_2017_CVPR} distance. We explore this opinion to cycle consistency loss by applying $L1$ distance to compute $L_{cyc}$. Our final objective is
\begin{align}
\label{eq:objective}
G^{*} = \arg\underset{G}{\min}\;\underset{D}{\max}\mathcal{L}(G,D)
+ \gamma \mathcal{L}_{cyc} + \sigma \mathcal{L}_{p}, \\
\text{w.r.t}\quad
\mathcal{L}(G,D)=
\begin{cases}
\mathcal{L}_{p}(G,D_{c}) \quad \text{for}\ D_{c},\\
\mathcal{L}_{g}(G,D_{f}) \quad \text{for}\ D_{f},
\end{cases}
\end{align}
where $\mathcal{L}_{cyc}$ means cycle consistency loss, $\mathcal{L}_{p}$ indicates perceptual loss, and $\gamma$ and $\sigma$ are hyper parameters to balance the contribution of each loss to the objective. We use greedy search to optimize the hyper parameters with $\gamma=10$ and $\sigma=2.0$ for all the experiments in this paper.

As shown in Figure~\ref{fig:p2c}, we use Conv-Residual blocks-Deconv~\cite{he2016deep} as the generator to share the low-level and high-level information between the input and output directly across the net.

\section{Experiments}

As a typical image-to-image translation task, photo-to-caricature requires high-level visual information conversion, which is very challenging for the state-of-the-art general-purpose solutions. To explore the effect of our proposed model, we tested the method on a variety of datasets for translating faces in the wild from photo to caricature with different styles, and the qualitative results are shown in Figs.~\ref{fig:comparison} and~\ref{fig:free}. 

\subsection{Dataset and training}

Our proposed model is trained in a supervised unpaired fashion on a paired face photo-caricature dataset, named IIIT-CFW-P2C dataset, which was rebuilt based on IIIT-CFW~\cite{mishra2016iiit}. The IIIT-CFW is a dataset for the cartoon faces in the wild and contains $8928$ annotated cartoon faces of famous personalities in the world with varying profession. Also it provides $1000$ real faces of the public figure for cross modal retrieval tasks. However, it's not suitable for the training of photo-to-caricature translation task using some paired methods (such as Pix2pix) because the face photos and face cartoons\footnote{We consider that caricature is a type of a cartoon or a subset of a cartoon.} are not paired, \emph{e.g.}, the facial orientation and expression of the photo and caricature for the same person are varying a lot. So we rebuild a photo-caricature dataset with $1171$ paired images by searching the IIIT-CFW dataset and Internet as the training set for compared experiments. Here we use $800$ for training and the left for testing. At inference time, we run the generator in exactly the same manner as during the training phase. Besides, we also extensively evaluated our method on a variety of datasets with faces in the wild, including Caricature~\cite{abaci2015matching}, FEI~\cite{thomaz2010new},  IIIT-CFW~\cite{mishra2016iiit}, Yale~\cite{georghiades1997yale}, KDEF~\cite{lundqvist1998karolinska} and CelebA~\cite{liu2015faceattributes}. 

Besides, we also consider photo-to-sketch as a photo-to-caricature task for experiments using PHOTO-SKETCH dataset~\cite{zhang2011coupled,wang2009face}, which has $1194$ paired images and hence can be directly used for supervised training of some compared paired methods. And following DualGAN, we use $995$ unpaired images for training and $199$ for testing. Note that we train CycleGAN, DualGAN and our model using unpaired images and train Pix2pix using paired images of the two datasets.

\subsection{Comparison with state-of-the-arts}

\begin{figure}[!ht]
\centering
\includegraphics[width=\textwidth]{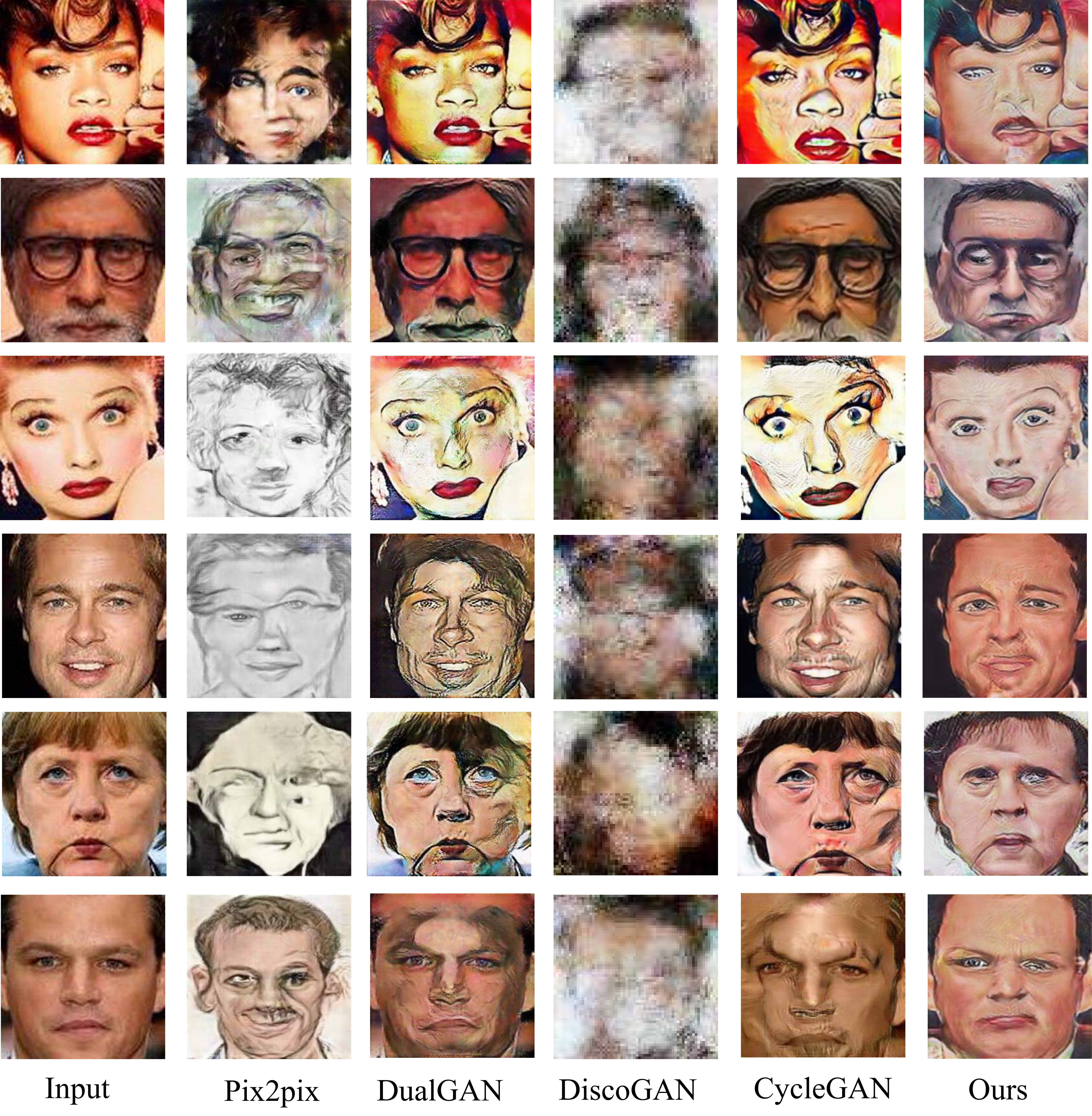}
\caption{Comparison of state-of-the-art image-to-image translation methods with our proposed method for face photo-to-caricature translation on IIIT-CFW-P2C dataset.}
\label{fig:comparison}
\end{figure}

Using IIIT-CFW-P2C dataset, we first compare our proposed method with Pix2pix~\cite{Isola_2017_CVPR}, DualGAN~\cite{Yi_2017_ICCV}, DiscoGAN~\cite{kim2017learning} and CycleGAN~\cite{Zhu_2017_ICCV} on photo-to-caricature translation task. All the four methods were trained on the same training dataset and tested on novel data from IIIT-CFW-P2C dataset that does not overlap those for training. 
\paragraph{Qualitative evaluation}
Figure~\ref{fig:comparison} shows the experimental results of the comparison, it can be seen that, DualGAN only learned the color and edge translation rather than structure information, Pix2pix makes structure error in almost all cases, while CycleGAN can keep the structure information of the input image but without enough conversion for caricature creation task. For DiscoGAN, it's really hard to generate plausible and meaningful results due to the lack data of training ($800$ pairs in our experiments vs. tens of thousands of pairs in DiscoGAN's experiments) and the big challenge of task (photo-to-caricature vs. edge-to-photo). Although it's still not good enough for the results of our method compared to human caricaturists, the experiments on photo-to-caricature translation of faces show considerable performance gain of our proposed method over state-of-the-art image-to-image translation methods, especially the encouraging ability of exaggeration and abstraction. However, due to the very challenging task with less training data but on various styles, our method also messes some details while translation (see the mouth of the first row and the eyes of the fourth row on our results in Figure~\ref{fig:comparison}).

This experiment also expresses that high-level image-to-image translation tasks like photo-to-caricature are generally more difficult than those low-level translation tasks such as photo-to-sketch, because it not only needs to abstract the facial features, but also requires to exaggerate the emotional expression. So that the pixel-level methods (like Pix2pix) might fail as they force the generator to concentrate on local information rather than whole structure.

\begin{figure}[!ht]
\centering
\includegraphics[width=\textwidth]{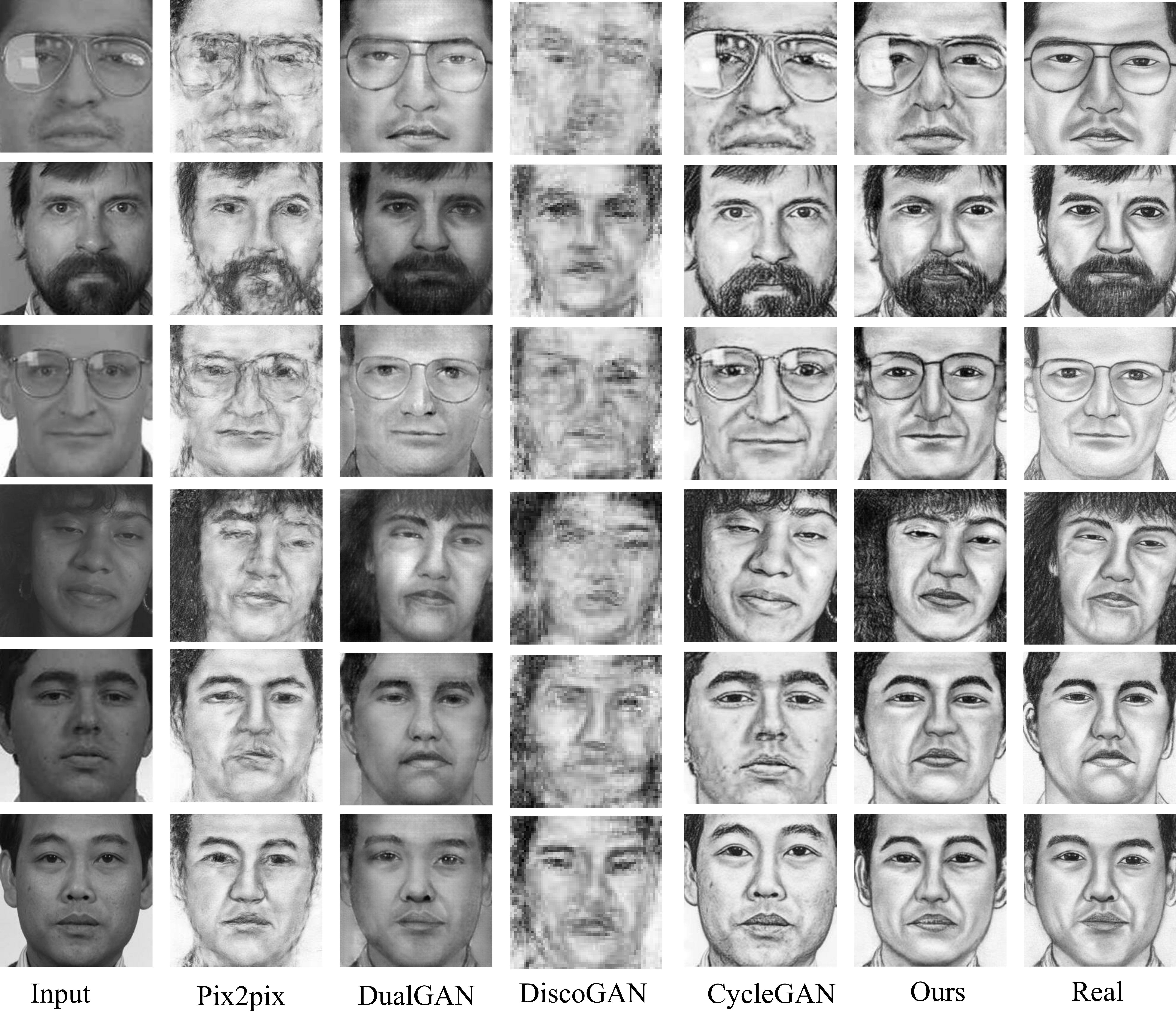}
\caption{Comparison of state-of-the-art image-to-image translation methods with our proposed method for face photo-to-caricature translation on PHOTO-SKETCH dataset.}
\label{fig:comparison_sketch}
\end{figure}

Besides, we also evaluate our methods on PHOTO-SKETCH dataset. Figure~\ref{fig:comparison_sketch} shows the compared results. Comparing with Pix2pix, our method can reduce the effect of being blurry and artifact. Although DualGAN and CycleGAN can also reserve the structure information of input faces, they are not good at achieving abstraction and artistry. Similarly, DiscoGAN collapses when facing insufficient training data and big challenging task.

\paragraph{Quantitative evaluation}
Beyond the visually qualitative evaluation, we also evaluated the translated cartoon results of different methods quantitatively on IIIT-CFW-P2C dataset and PHOTO-SKETCH dataset in terms of both human judge and machine grade, and the average results are shown in Table~\ref{tab:cfw} and Table~\ref{tab:photo} respectively.

For human score, we invited 40 volunteers to evaluate the generated image quality of different methods in terms of satire, exaggeration, lifelikeness and artistry compared with the given original photo, by grading from 1 to 5 (1 represents the worst and 5 represents the best). 
For inception score, we used a pre-trained classifier network and sampled images for evaluation followed~\cite{salimans2016improved}. The results shows that our proposed method outperforms state-of-the-art image-to-image translation methods with the highest human and inception scores.
\begin{table}
\centering
\caption{The generated image equality evaluation results of different methods on IIIT-CFW-P2C dataset.}
\label{tab:cfw}
\begin{tabular}{|c|c|c|}
\hline
Method & Human score & Inception score \\
\hline\hline
Pix2pix & 2.0106 & 1.5069$\pm$0.1090\\
DualGAN & 1.9946 & 1.4843$\pm$0.1049\\
DiscoGAN & 1.5014 & 1.3366$\pm$0.0714\\
CycleGAN & 3.5001 & 1.5684$\pm$0.1331\\
Ours & 4.0120 & 1.6043$\pm$0.0918 \\
\hline
\end{tabular}
\end{table}
\begin{table}
\centering
\caption{The generated image equality evaluation results of different methods on PHOTO-SKETCH dataset.}
\label{tab:photo}
\begin{tabular}{|c|c|c|}
\hline
Method & AMT & Inception score \\
\hline\hline
Pix2pix & 2.0971 & 1.3625$\pm$0.0706\\
DualGAN & 2.3810 & 1.4063$\pm$0.0822\\
DiscoGAN & 1.0858 & 1.3142$\pm$0.0206\\
CycleGAN & 3.2272 & 1.3980$\pm$0.1130\\
Ours & 4.0750 & 1.4298$\pm$0.0818 \\
\hline
\end{tabular}
\end{table}

\subsection{Dual discriminators}
\label{sec:dual}
\begin{figure}[!ht]
\centering
\includegraphics[width=0.5\textwidth]{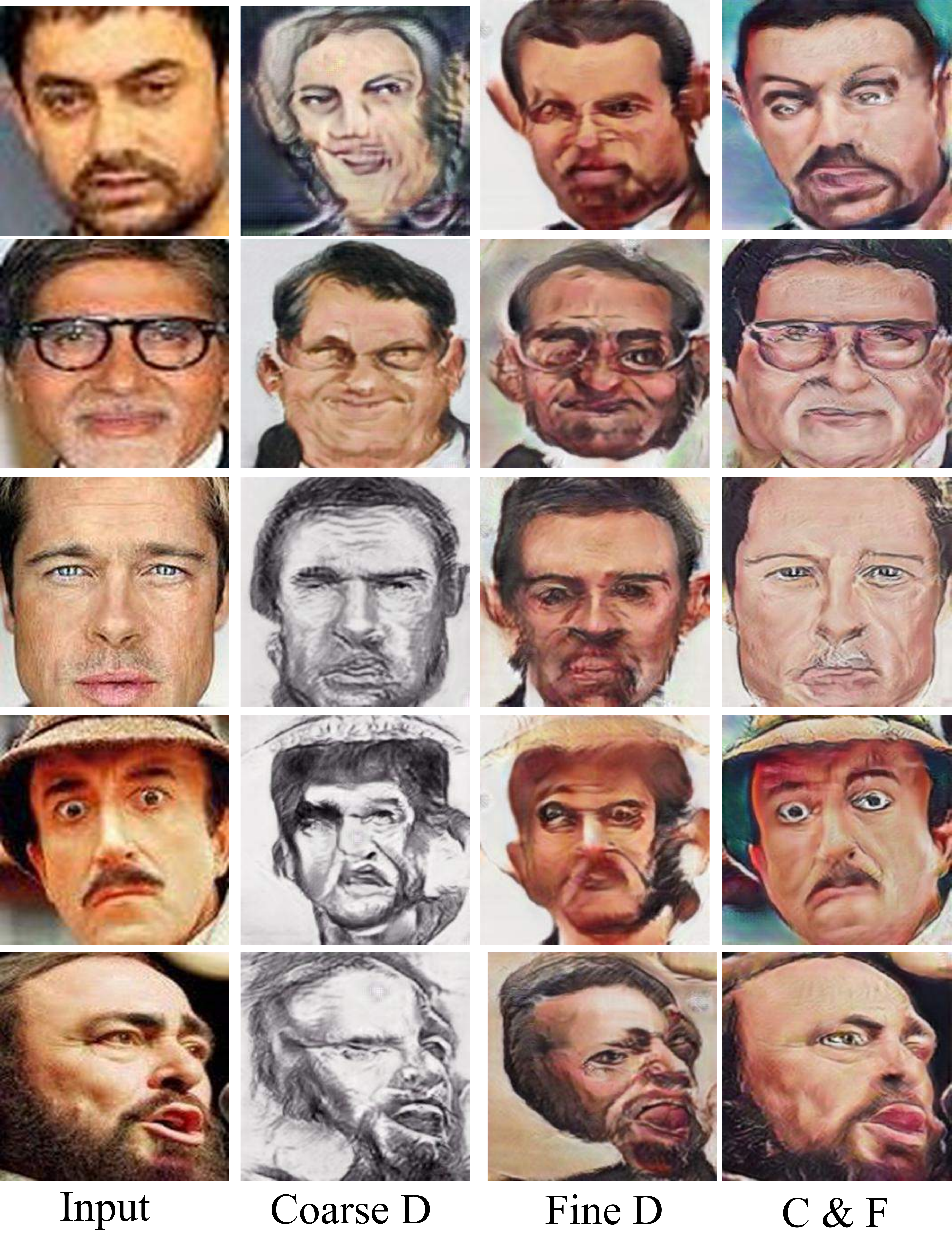}
\caption{Comparison of only one discriminator (Coarse D and Fine D respectively) with our dual discriminators (Coarse D + Fine D, C \& F).}
\label{fig:dual}
\end{figure}

In this experiment, we verify the effectiveness of our proposed dual pathway of discriminators. We first use only one coarse discriminator (Coarse D) and one fine discriminator (Fine D) separately, and then use dual discriminators with one coarse discriminator plus one fine discriminator (Fine D + Coarse D), while keeping all other architectures and settings fixed for training and testing. Some example results are shown in Figure~\ref{fig:dual}, and one Fine D model almost misses the key structure information on faces, but our dual Coarse D + Fine D model can render the structure of facial features well. It further proves that the Fine D model only concerns the local statistics for tackling low-level image-to-image translation tasks (\emph{e.g.}, photo-to-sketch).

\begin{figure}[!ht]
\centering
\includegraphics[width=\textwidth]{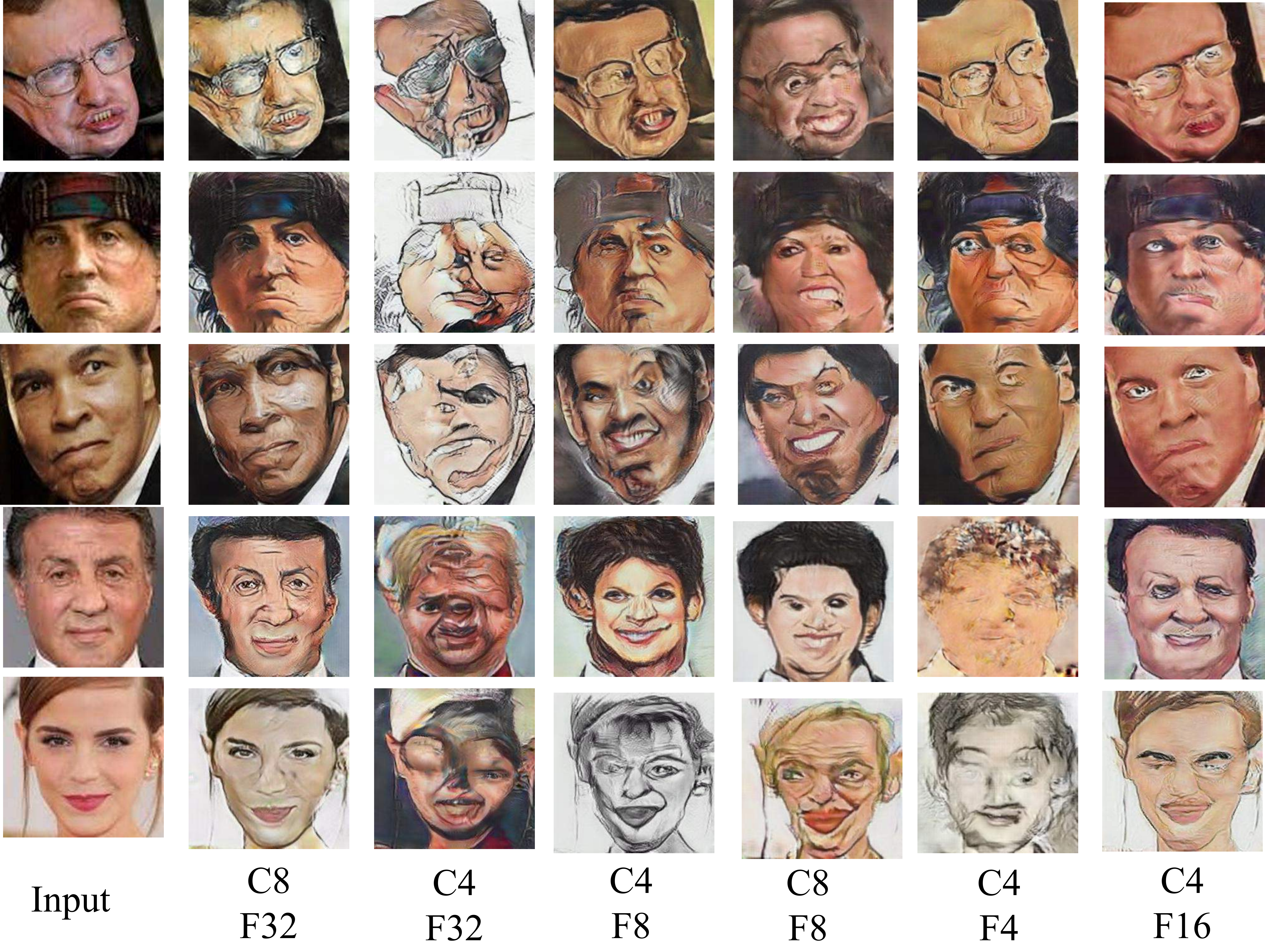}
\caption{Examples of different combination sizes of output patches for coarse discriminator and fine discriminator. C$k$ denotes that the output of coarse discriminator is a $k\times k$ patch, and F$k$ denotes that the output of fine discriminator is a $k\times k$ patch. It can be seen that large F$k$ such as F$32$ for fine discriminator fails to abstract faces and achieve exaggerations, while small F$k$ such as F$8$ abstracts and exaggerates faces too much.}
\label{fig:combination}
\end{figure}

And we also took some experiments to greedy search the best combination size of output patches for coarse discriminator and fine discriminator. Figure~\ref{fig:combination} shows results of different combinations, which indicates that large Fine D patches (\emph{e.g.}, F$32$) fails to abstract and exaggerate faces while small Fine D patches (\emph{e.g.}, F$8$) abstracts and exaggerates faces too much. 

\subsection{Loss selection}
\label{sec:loss}

\begin{figure}[!ht]
\centering
\includegraphics[width=0.5\textwidth]{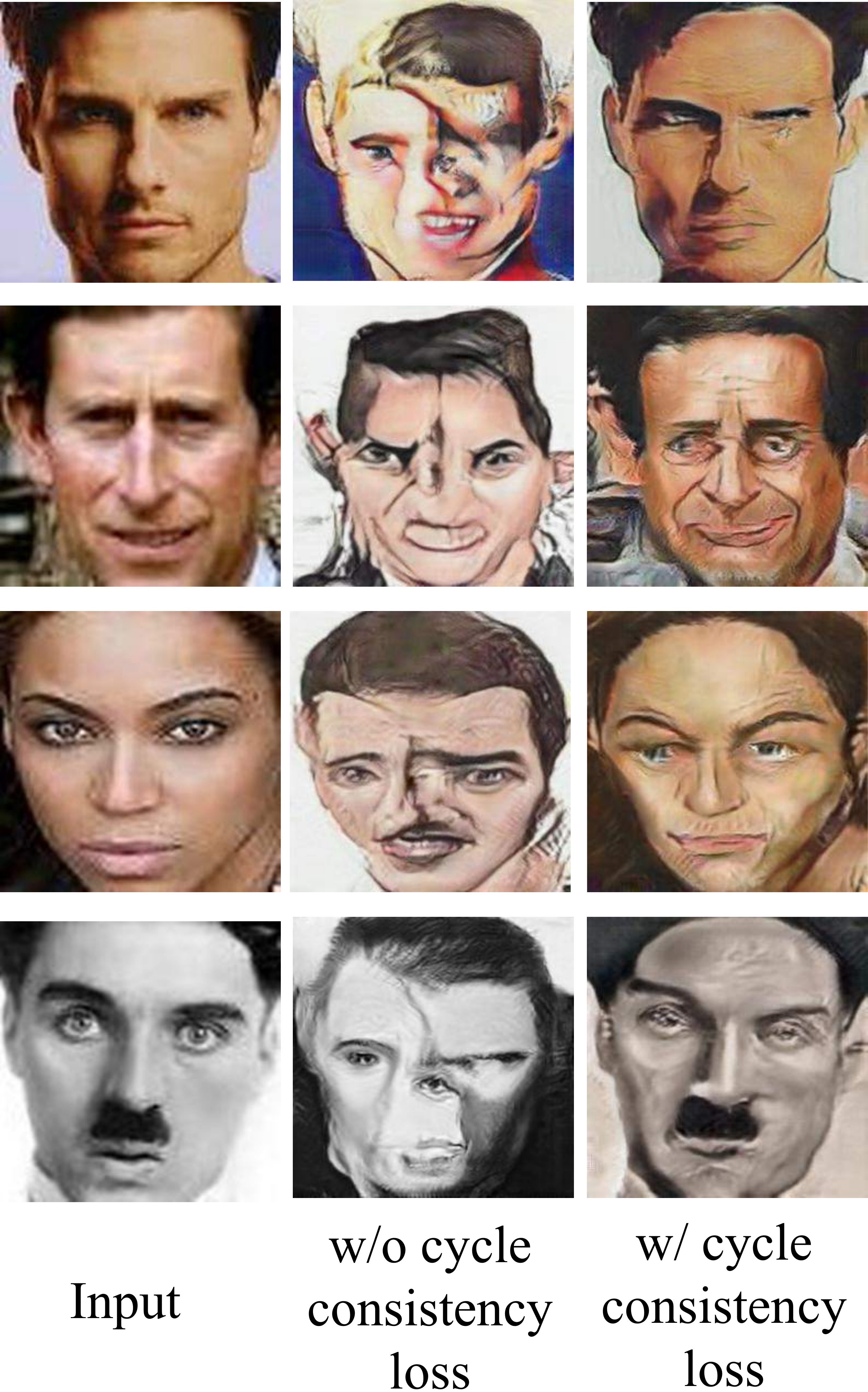}
\caption{Comparison of extra loss for final objective of generator: without ($w/o$) $\mathcal{L}_{cyc}$ and with ($w/$) $\mathcal{L}_{cyc}$.}
\label{fig:losscyc}
\end{figure}

We first consider to check if the cycle consistency loss $\mathcal{L}_{cyc}$ should be provided to the GAN objective (Equation~\ref{eq:objective}), and the second column in Figure~\ref{fig:losscyc} shows the results without cycle consistency loss. It's easy to see that without cycle consistency loss, although the adversarial system with adversarial loss can capture the facial features, it's hard to generate caricature images with plausible objects and meaningful relationship between facial organs. The third column in Figure~\ref{fig:losscyc} shows the results by adding cycle consistency loss $\mathcal{L}_{cyc}$. Therefore, the normal adversarial training can lead to some kind of caricature style, but it fails to be lifelike without meaningful components.

\begin{figure}[t]
\centering
\includegraphics[width=0.5\linewidth]{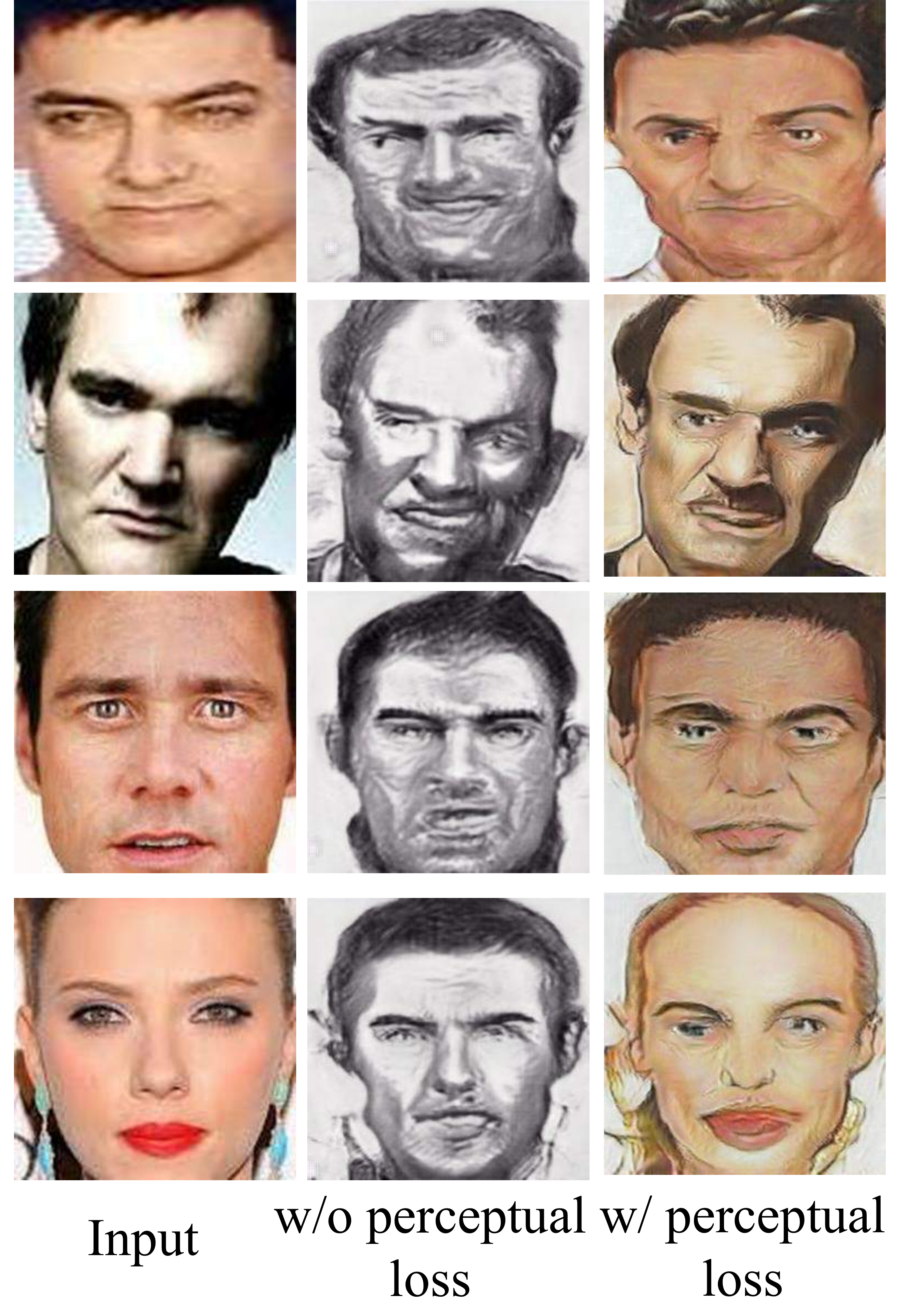}
\caption{Comparison of extra loss for final objective of generator: without ($w/o$) $\mathcal{L}_{p}$ and with ($w/$) $\mathcal{L}_{p}$.}
\label{fig:losscl}
\end{figure}

Based on the cycle consistency loss, we provide the perceptual loss $\mathcal{L}_{p}$ for generator in our system. Figure~\ref{fig:losscl} shows the compared results of without and with $\mathcal{L}_{p}$. It can be seen that perceptual loss can produce images with the exaggerated facial features such as eyes, nose, and mouth. The perceptual loss, which expresses some perceptual errors on facial features such as head, eyes, and mouth, could improve the artistic expression of image generation and show better abstraction ability. And it can also reduce the effect of being blurry. The second column of Figure~\ref{fig:losscl} without using perceptual loss illustrates the indistinguishable facial expressions with distorted facial organs while translation, and the third column with adding perceptual loss improves the performance on facial expression and organ translation with caricature effect, \emph{e.g.}, the smile woman of last row.

\subsection{Auxiliary noise input}
\label{sec:info}
\begin{figure}[!ht]
\centering
\includegraphics[width=\textwidth]{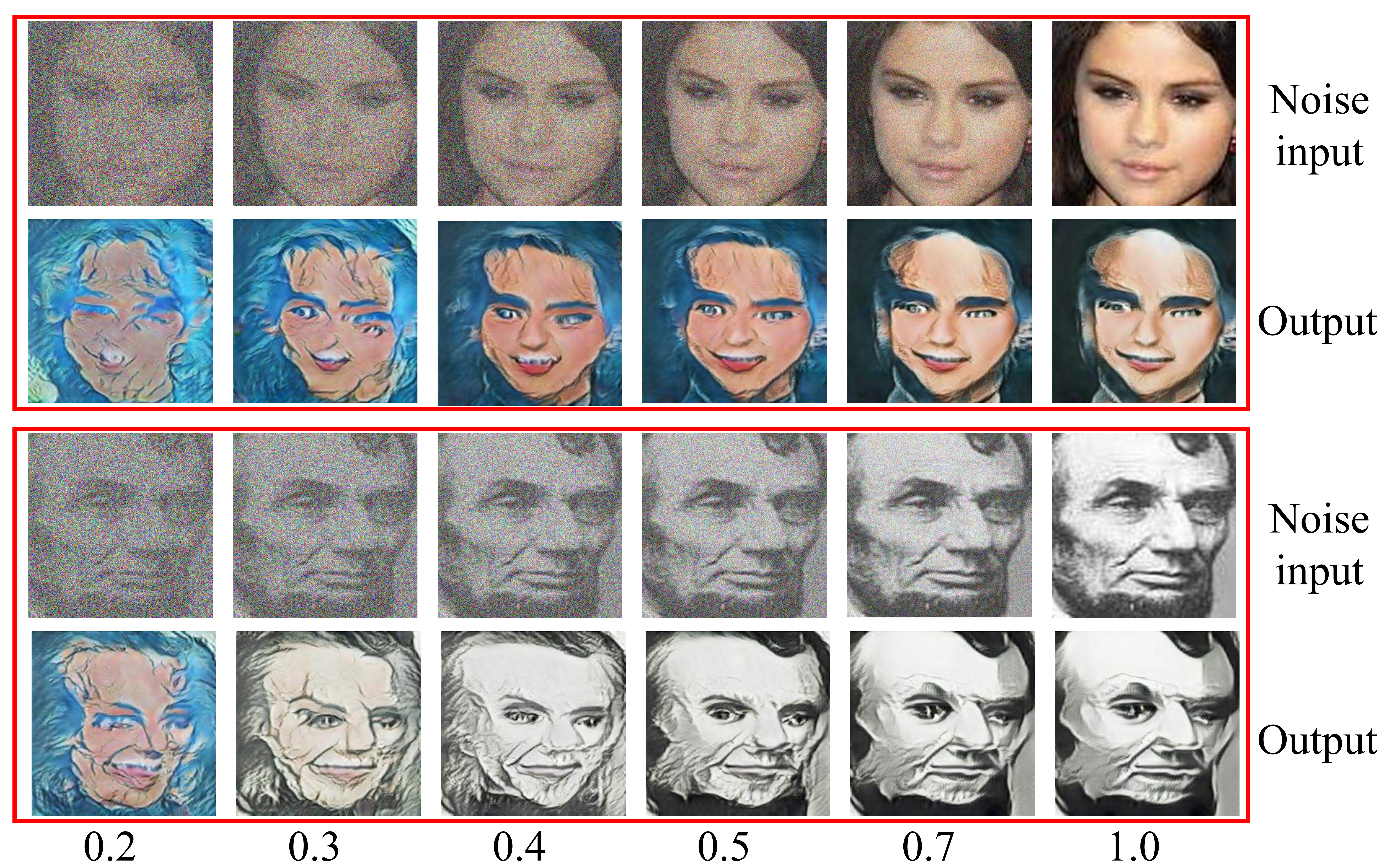}
\caption{Examples of adding auxiliary noise for robustness and diversity of our photo-to-caricature translation. Different proportions ($\alpha$ in Equation~\ref{fig:noise}) of noise inputs can also lead to meaningful different styles of output caricatures.}
\label{fig:info}
\end{figure}

By adding auxiliary noise to our photo-to-caricature system, we can improve the robustness and diversity of synthesized facial caricatures. Figure~\ref{fig:info} shows the example results of adding auxiliary noise for photo-to-caricature translation, and the output results for adding different proportions ($\alpha$) of noise indicate that our system can still synthesize meaningful facial caricatures with even more than a half noise ($1-\alpha$, see Equation~\ref{eq:noise} for reference) as inputs, besides, the added different proportions of noise also lead to different styles of output results, which indicates that it might be used as a factor for tuning different synthesized styles.

\subsection{Freestyle face caricature creation}

\begin{figure}[!ht]
\centering
\includegraphics[width=\linewidth]{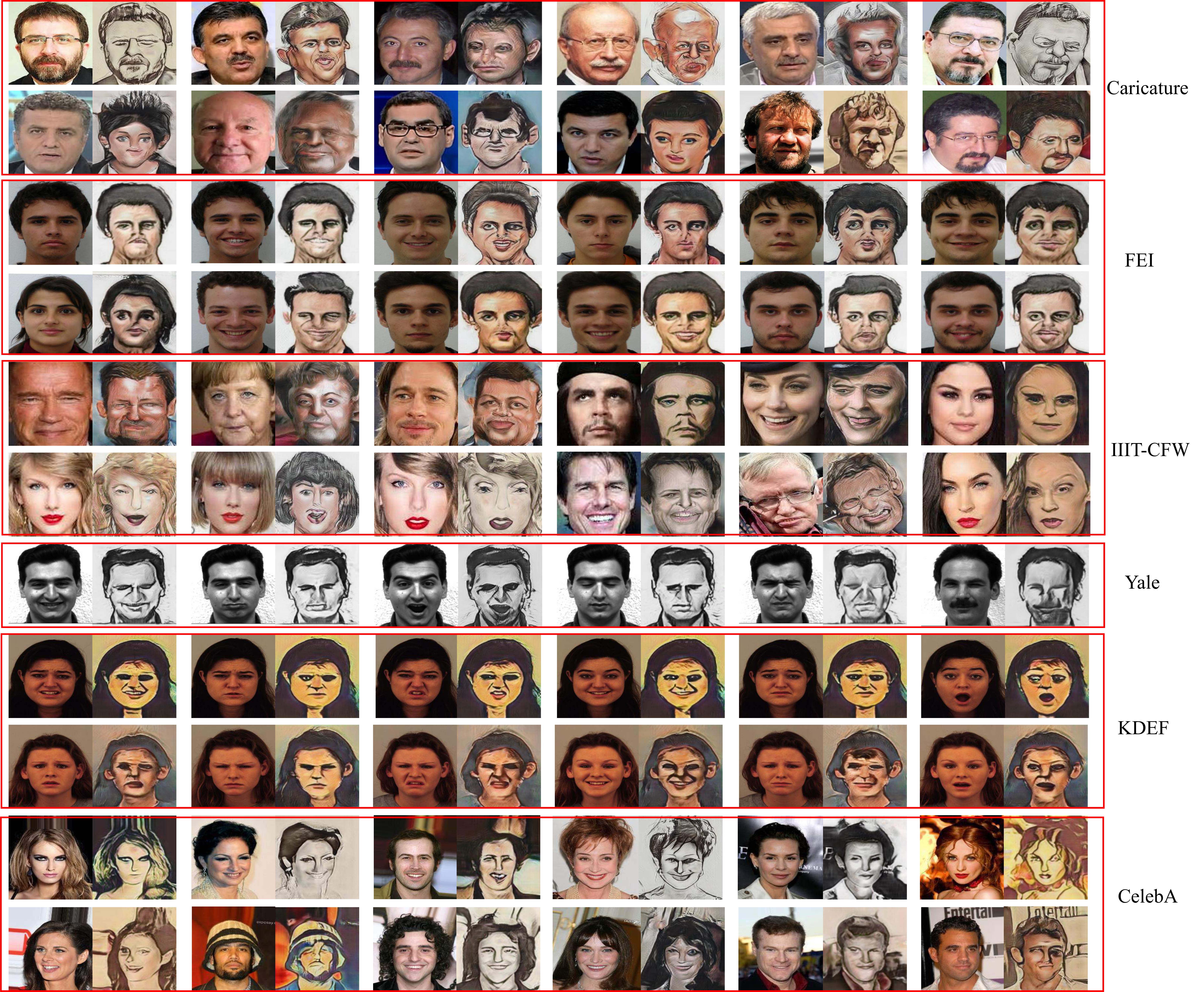}
\caption{Example translated caricatures of facial photos from several datasets (Caricature~\cite{abaci2015matching}, FEI~\cite{thomaz2010new},  IIIT-CFW~\cite{mishra2016iiit}, Yale~\cite{georghiades1997yale}, KDEF~\cite{lundqvist1998karolinska} and CelebA~\cite{liu2015faceattributes}) using our trained model on IIIT-CFW-P2C.}
\label{fig:free}
\end{figure}

To evaluate the ability for real applications in the daily life, we tested our method on a variety of face datasets, including Caricature~\cite{abaci2015matching},  FEI~\cite{thomaz2010new}, IIIT-CFW~\cite{mishra2016iiit}, Yale~\cite{georghiades1997yale}, KDEF~\cite{lundqvist1998karolinska} and CelebA~\cite{liu2015faceattributes}, to illustrate the photo-to-caricature translation on faces in the wild, and the results are shown in Figure~\ref{fig:free}. These freestyle face caricature creation results validate that our model works not bad on arbitrary faces and show the potential value for the related applications. And we can see that the translated results in KDEF, FEI and Yale datasets also have different facial expression corresponding the input faces. Our methods successfully reserve the emotion information and emulate the facial organs with caricature style. So we can conclude that the more abstracted information such as facial emotion and expression with global structure information are reserved. Besides, our model can also enlarge or narrow the facial organs such as chin, lips, eyes and so on, which is required for high-level image-to-image translation tasks. 

\section{Conclusion and Future Work}

We present a novel GAN-based method to deal with high-level image-to-image translation task, \emph{i.e.}, photo-to-caricature translation. The proposed method uses dual discriminators for capturing global structure and local statistics information with abstraction ability, also provides extra perceptual loss on GAN objective to constrain the consistency under exaggeration. Besides, the style information can be learned and representative by adding auxiliary noise input. And the robustness can be improved by the noise-added training. Experimental results show that our method not only outperforms other state-of-the-art image-to-image translation methods, but also works well on a variety of datasets for photo-to-caricature translation of faces in the wild.

\textbf{Limitations.} Translating photo to caricature is a very challenging high-level image-to-image translation task. Thus, our model also fails in some cases, \emph{e.g.}, some generated images of Yale and CelebA dataset in Figure~\ref{fig:free}. Although our method can keep the structure information of faces, it is still hard to render the details for providing high-quality caricatures and some tiny organs (such as eyes) are lack of details in Figure~\ref{fig:comparison}. Besides, it's also sensitive to side face with complex background, \emph{e.g.}, some cases on CelebA and KDEF datasets in Figure~\ref{fig:free}.

\textbf{Future work.} With regard to future work, first, it would be interesting to investigate our method on other tasks of high-level image-to-image translation (\emph{e.g.}, human-to-cartoon translation for cartoon movies); second, for the proposed method, the model still needs to be improved to provide high-quality rendered translated results; third, we intend to apply our method on the real applications of automatic and intelligent photo-to-caricature translation; fourth, we hope that we can control caricature style while translating images between domains by tuning input noise.

\section*{Acknowledgement}
We thanks the volunteers for grading human scores of translation results from different methods. This work was supported by the National Natural Science Foundation of China [61771440, 41776113], and Qingdao Municipal Science and Technology Program [17-1-1-5-jch].


\bibliography{P2C}

\begin{thebibliography}{10}
\expandafter\ifx\csname url\endcsname\relax
  \def\url#1{\texttt{#1}}\fi
\expandafter\ifx\csname urlprefix\endcsname\relax\def\urlprefix{URL }\fi
\expandafter\ifx\csname href\endcsname\relax
  \def\href#1#2{#2} \def\path#1{#1}\fi

\bibitem{Isola_2017_CVPR}
P.~Isola, J.-Y. Zhu, T.~Zhou, A.~A. Efros, Image-to-image translation with
  conditional adversarial networks, in: CVPR, 2017.

\bibitem{Zhu_2017_ICCV}
J.-Y. Zhu, T.~Park, P.~Isola, A.~A. Efros, Unpaired image-to-image translation
  using cycle-consistent adversarial networks, in: ICCV, 2017.

\bibitem{Yi_2017_ICCV}
Z.~Yi, H.~Zhang, P.~Tan, M.~Gong, {DualGAN}: Unsupervised dual learning for
  image-to-image translation, in: ICCV, 2017.

\bibitem{Chen_2017_ICCV}
Q.~Chen, V.~Koltun, Photographic image synthesis with cascaded refinement
  networks, in: ICCV, 2017.

\bibitem{kim2017learning}
T.~Kim, M.~Cha, H.~Kim, J.~Lee, J.~Kim, Learning to discover cross-domain
  relations with generative adversarial networks, arXiv preprint
  arXiv:1703.05192.

\bibitem{mao2018semantic}
X.~Mao, S.~Wang, L.~Zheng, Q.~Huang, Semantic invariant cross-domain image
  generation with generative adversarial networks, Neurocomputing 293 (2018)
  55--63.

\bibitem{zhou2016similarity}
Y.~Zhou, X.~Bai, W.~Liu, L.~J. Latecki, Similarity fusion for visual tracking,
  International Journal of Computer Vision 118~(3) (2016) 337--363.

\bibitem{wang2016super}
Q.~Wang, S.~Li, H.~Qin, A.~Hao, Super-resolution of multi-observed {RGB-D}
  images based on nonlocal regression and total variation, IEEE Transactions on
  Image Processing 25~(3) (2016) 1425--1440.

\bibitem{ma2017unsupervised}
J.~Ma, S.~Li, H.~Qin, A.~Hao, Unsupervised multi-class co-segmentation via
  joint-cut over $l_1$-manifold hyper-graph of discriminative image regions,
  IEEE Transactions on Image Processing 26~(3) (2017) 1216--1230.

\bibitem{shi2017end}
B.~Shi, X.~Bai, C.~Yao, An end-to-end trainable neural network for image-based
  sequence recognition and its application to scene text recognition, IEEE
  Transactions on Pattern Analysis and Machine Intelligence 39~(11) (2017)
  2298--2304.

\bibitem{zhu2016generative}
J.-Y. Zhu, P.~Kr{\"a}henb{\"u}hl, E.~Shechtman, A.~A. Efros, Generative visual
  manipulation on the natural image manifold, in: ECCV, 2016.

\bibitem{li2016precomputed}
C.~Li, M.~Wand, Precomputed real-time texture synthesis with {M}arkovian
  generative adversarial networks, in: ECCV, 2016.

\bibitem{wang2016generative}
X.~Wang, A.~Gupta, Generative image modeling using style and structure
  adversarial networks, in: ECCV, 2016.

\bibitem{Ledig_2017_CVPR}
C.~Ledig, L.~Theis, F.~Huszar, J.~Caballero, A.~Cunningham, A.~Acosta,
  A.~Aitken, A.~Tejani, J.~Totz, Z.~Wang, W.~Shi, Photo-realistic single image
  super-resolution using a generative adversarial network, in: CVPR, 2017.

\bibitem{taigman2017unsupervised}
Y.~Taigman, A.~Polyak, L.~Wolf, Unsupervised cross-domain image generation, in:
  ICLR, 2017.

\bibitem{Sela_2017_ICCV}
M.~Sela, E.~Richardson, R.~Kimmel, Unrestricted facial geometry reconstruction
  using image-to-image translation, in: ICCV, 2017.

\bibitem{Tung_2017_ICCV}
H.-Y. Fish~Tung, A.~W. Harley, W.~Seto, K.~Fragkiadaki, Adversarial inverse
  graphics networks: Learning {2D}-to-{3D} lifting and image-to-image
  translation from unpaired supervision, in: ICCV, 2017.

\bibitem{goodfellow2014generative}
I.~Goodfellow, J.~Pouget-Abadie, M.~Mirza, B.~Xu, D.~Warde-Farley, S.~Ozair,
  A.~Courville, Y.~Bengio, Generative adversarial nets, in: NIPS, 2014.

\bibitem{mirza2014conditional}
M.~Mirza, S.~Osindero, Conditional generative adversarial nets, arXiv preprint
  arXiv:1411.1784.

\bibitem{radford2016unsupervised}
A.~Radford, L.~Metz, S.~Chintala, Unsupervised representation learning with
  deep convolutional generative adversarial networks, in: ICLR, 2016.

\bibitem{liu2018auto}
Y.~Liu, Z.~Qin, T.~Wan, Z.~Luo, Auto-painter: Cartoon image generation from
  sketch by using conditional wasserstein generative adversarial networks,
  Neurocomputing.

\bibitem{akleman2000making}
E.~Akleman, J.~Palmer, R.~Logan, Making extreme caricatures with a new
  interactive {2D} deformation technique with simplicial complexes, in:
  Proceedings of Visual, 2000, pp. 100--105.

\bibitem{radford2015unsupervised}
A.~Radford, L.~Metz, S.~Chintala, Unsupervised representation learning with
  deep convolutional generative adversarial networks, arXiv preprint
  arXiv:1511.06434.

\bibitem{berthelot2017began}
D.~Berthelot, T.~Schumm, L.~Metz, {BEGAN}: {B}oundary {E}quilibrium
  {G}enerative {A}dversarial {N}etworks, arXiv preprint arXiv:1703.10717.

\bibitem{mishra2016iiit}
A.~Mishra, S.~N. Rai, A.~Mishra, C.~Jawahar, {IIIT-CFW}: A benchmark database
  of cartoon faces in the wild, in: ECCVW, 2016.

\bibitem{zhang2011coupled}
W.~Zhang, X.~Wang, X.~Tang, Coupled information-theoretic encoding for face
  photo-sketch recognition, in: CVPR, IEEE, 2011.

\bibitem{wang2009face}
X.~Wang, X.~Tang, Face photo-sketch synthesis and recognition, IEEE
  Transactions on Pattern Analysis and Machine Intelligence 31~(11) (2009)
  1955--1967.

\bibitem{chen2016infogan}
X.~Chen, Y.~Duan, R.~Houthooft, J.~Schulman, I.~Sutskever, P.~Abbeel,
  Info{GAN}: Interpretable representation learning by information maximizing
  generative adversarial nets, in: NIPS, 2016.

\bibitem{abaci2015matching}
B.~Abaci, T.~Akgul, Matching caricatures to photographs, Signal, Image and
  Video Processing 9~(1) (2015) 295--303.

\bibitem{thomaz2010new}
C.~E. Thomaz, G.~A. Giraldi, A new ranking method for principal components
  analysis and its application to face image analysis, Image and Vision
  Computing 28~(6) (2010) 902--913.

\bibitem{georghiades1997yale}
A.~Georghiades, P.~Belhumeur, D.~Kriegman, Yale face database, Center for
  computational Vision and Control at Yale University,
  http://cvc.cs.yale.edu/cvc/projects/yalefaces/yalefaces.html 2.

\bibitem{lundqvist1998karolinska}
D.~Lundqvist, A.~Flykt, A.~{\"O}hman, The karolinska directed emotional faces
  ({KDEF}), CD ROM from Department of Clinical Neuroscience, Psychology
  Section, Karolinska Institutet.

\bibitem{liu2015faceattributes}
Z.~Liu, P.~Luo, X.~Wang, X.~Tang, Deep learning face attributes in the wild,
  in: ICCV, 2015.

\bibitem{mathieu2015deep}
M.~Mathieu, C.~Couprie, Y.~LeCun, Deep multi-scale video prediction beyond mean
  square error, arXiv preprint arXiv:1511.05440.

\bibitem{reed2016generative}
S.~Reed, Z.~Akata, X.~Yan, L.~Logeswaran, B.~Schiele, H.~Lee, Generative
  adversarial text to image synthesis, in: ICML, 2016.

\bibitem{yan2016attribute2image}
X.~Yan, J.~Yang, K.~Sohn, H.~Lee, Attribute2image: Conditional image generation
  from visual attributes, in: ECCV, 2016.

\bibitem{zhang2017image}
H.~Zhang, V.~Sindagi, V.~M. Patel, Image de-raining using a conditional
  generative adversarial network, arXiv preprint arXiv:1701.05957.

\bibitem{Pathak_2016_CVPR}
D.~Pathak, P.~Krahenbuhl, J.~Donahue, T.~Darrell, A.~A. Efros, Context
  encoders: Feature learning by inpainting, in: CVPR, 2016.

\bibitem{iizuka2017globally}
S.~Iizuka, E.~Simo-Serra, H.~Ishikawa, Globally and locally consistent image
  completion, ACM Transactions on Graphics 36~(4) (2017) 1--14.

\bibitem{luo2002exaggeration}
W.~C. Luo, P.~C. Liu, M.~Ouhyoung, Exaggeration of facial features in
  caricaturing, in: Proceedings of International Computer Symposium, 2002.

\bibitem{chen2004automatic}
Y.-L. Chen, W.-H. Tsai, et~al., Automatic generation of talking cartoon faces
  from image sequences, in: Proceedings of Conferences on Computer Vision,
  Graphics \& Image Processing, 2004.

\bibitem{liao2004automatic}
P.-Y. C. W.-H. Liao, T.-Y. Li, Automatic caricature generation by analyzing
  facial features, in: ACCV, 2004.

\bibitem{zhang2017style}
L.~Zhang, Y.~Ji, X.~Lin, Style transfer for anime sketches with enhanced
  residual {U}-net and auxiliary classifier {GAN}, arXiv preprint
  arXiv:1706.03319.

\bibitem{gatys2016image}
L.~A. Gatys, A.~S. Ecker, M.~Bethge, Image style transfer using convolutional
  neural networks, in: CVPR, IEEE, 2016.

\bibitem{bruna2015super}
J.~Bruna, P.~Sprechmann, Y.~LeCun, Super-resolution with deep convolutional
  sufficient statistics, arXiv preprint arXiv:1511.05666.

\bibitem{dosovitskiy2016generating}
A.~Dosovitskiy, T.~Brox, Generating images with perceptual similarity metrics
  based on deep networks, in: NIPS, 2016.

\bibitem{johnson2016perceptual}
J.~Johnson, A.~Alahi, L.~Fei-Fei, Perceptual losses for real-time style
  transfer and super-resolution, in: ECCV, Springer, 2016.

\bibitem{ledig2016photo}
C.~Ledig, L.~Theis, F.~Husz{\'a}r, J.~Caballero, A.~Cunningham, A.~Acosta,
  A.~Aitken, A.~Tejani, J.~Totz, Z.~Wang, et~al., Photo-realistic single image
  super-resolution using a generative adversarial network, arXiv preprint
  arXiv:1609.04802.

\bibitem{zhao2016energy}
J.~Zhao, M.~Mathieu, Y.~LeCun, Energy-based generative adversarial network,
  arXiv preprint arXiv:1609.03126.

\bibitem{he2016deep}
K.~He, X.~Zhang, S.~Ren, J.~Sun, Deep residual learning for image recognition,
  in: CVPR, 2016.

\bibitem{salimans2016improved}
T.~Salimans, I.~Goodfellow, W.~Zaremba, V.~Cheung, A.~Radford, X.~Chen,
  Improved techniques for training {GANs}, in: NIPS, 2016.

\end{thebibliography}

\end{document}